\documentclass{article}
\usepackage{algorithm}
\usepackage{algorithmic}
\usepackage{algorithm}
\usepackage{algorithmic}
\usepackage{amsmath}
\usepackage{booktabs} 
\usepackage{multirow} 
\usepackage{amsmath}
\usepackage{amsfonts}
\usepackage{spconf,amsmath,graphicx,hyperref}
\usepackage{subcaption} 

\title{Adaptive Reasoning Executor: A Collaborative Agent System for Efficient Reasoning}
\name{Zehui Ling$^{1,3}$ \  Deshu Chen$^{1,2,3}$ \ Yichi Zhang$^{1,2,3}$ \ Yuchen Liu$^{1,3}$  \ Xigui Li$^{1,3}$ \ Xin Guo$^{1,3}$ \ Yuan Cheng$^{1,3,\dagger}$ }
  \address{$^{1}$ Artificial Intelligence Innovation and Incubation Institute, Fudan University \\
      $^{2}$ School of Data Science, Fudan University\\
      $^{3}$ Shanghai Academy of Artificial Intelligence for Science}
\begin{document}
%
\maketitle
\begin{abstract}
Recent advances in Large Language Models (LLMs) demonstrate that chain-of-thought prompting and deep reasoning substantially enhance performance on complex tasks, and multi-agent systems can further improve accuracy by enabling model debates. However, applying deep reasoning to all problems is computationally expensive. To mitigate these costs, we propose a complementary agent system integrating small and large LLMs. The small LLM first generates an initial answer, which is then verified by the large LLM. If correct, the answer is adopted directly; otherwise, the large LLM performs in-depth reasoning. Experimental results show that, for simple problems, our approach reduces the computational cost of the large LLM by more than 50\% with negligible accuracy loss, while consistently maintaining robust performance on complex tasks.

\end{abstract}
\begin{keywords}
Large Language Model, Efficient Reasoning, Agent System
\end{keywords}
\section{Introduction}
\label{sec:intro}
In recent years, Large Language Models (LLMs) have advanced rapidly. Recent research has proposed reasoning strategies such as Chain of Thought (CoT) \cite{wei2022chain}, which guides the model to reason step by step by adding prompts such as “Let’s think step by step”. Multiple studies \cite{kojima2022large,yao2023tree,wang2022self} have demonstrated that this strategy can significantly enhance the quality of model responses. Building on this insight, many state-of-the-art models, such as DeepSeek \cite{guo2025deepseek}, Gemini \cite{comanici2025gemini}, and GPT \cite{agarwal2025gpt}, now incorporate deep reasoning capabilities. Beyond single-model approaches, multi-agent debate frameworks have been introduced, aiming to leverage interactions among multiple agents to strengthen the capability of tackling complex problem-solving tasks \cite{du2023improving,zhang2025stop}. 

While these approaches aim to solve difficult problems through more elaborate reasoning, applying deep reasoning indiscriminately to all tasks incurs substantial computational and inference costs. Even for simple arithmetic questions, forcing the model to reason extensively can result in unnecessary token overhead. This naturally raises the question: \textit{Can we adopt differentiated reasoning strategies based on task difficulty?} For simple problems, the model can directly produce answers without unnecessary reasoning, whereas for more challenging tasks, deep reasoning or multi-agent collaboration can be employed to ensure accuracy. To this end, several complementary approaches have been proposed \cite{sui2025stop,zhao2025let}. TokenSkip \cite{xia2025tokenskip} assigns an importance score to each token in a response and prunes those deemed uninformative, thereby producing more efficient training data. ThinkPrune \cite{hou2025thinkprune} leverages reinforcement learning by rewarding correct answers more strongly than incorrect ones, while progressively constraining the maximum generation length during training, ultimately encouraging the model to produce concise yet accurate reasoning. While single-model approaches are effective in producing more concise responses, they often demand substantial computational resources during training.

\begin{figure*}[htbp]
\centering
    \begin{subfigure}[b]{2\columnwidth}
        \includegraphics[width=\linewidth]{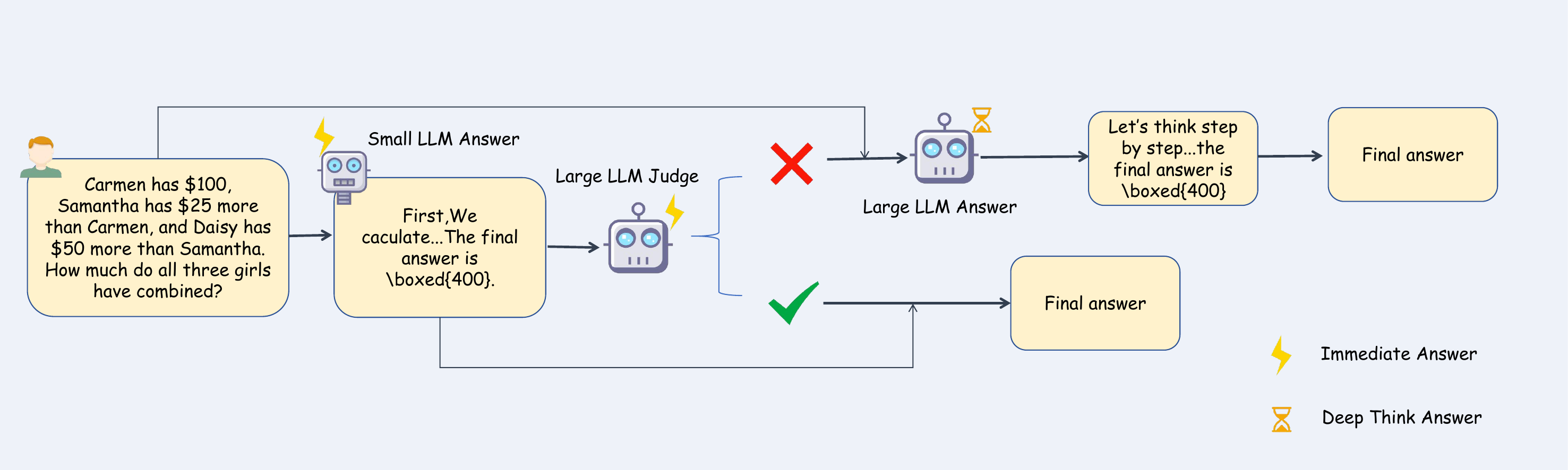}
    \end{subfigure}
    \caption{In our pipeline, the small LLM first generates an initial answer, which is then verified by the large LLM. If the initial answer is judged to be correct, it is adopted directly. Otherwise, the large LLM conducts in-depth reasoning to produce the final result.}
    \label{fig:1}
\end{figure*}

In this work, we propose an efficient agent system that integrates both small and large LLMs in a complementary manner. Specifically, the small LLM acts as an initial agent, providing a preliminary assessment of the problem, after which the problem and its answer are evaluated by the large LLM agent. If the large LLM agent deems the evaluation satisfactory, the answer is directly accepted; otherwise, the large LLM agent engages in deeper reasoning to deliver the final decision. Experimental results demonstrate that this agent-based approach can reduce nearly half of the large LLM’s computational cost for simple problems, while incurring only a minimal loss in accuracy. Furthermore, for more challenging problems, the model’s performance remains largely unchanged. As a result, the computational cost associated with invoking the large LLM API is greatly reduced.
Our contributions are organized as follows: 1) 
We propose a pipeline that determines whether the answer of small LLM can be directly adopted based on large LLM’s assessment. 2) We introduce two evaluation strategies, one that directly assesses the correctness of the answer and another that performs step-by-step verification. 3) We empirically validate the effectiveness of the proposed approach through experiments.

\section{Method}
\begin{figure}[h]
\centering
    \begin{subfigure}[b]{1\columnwidth}
        \includegraphics[width=\linewidth]{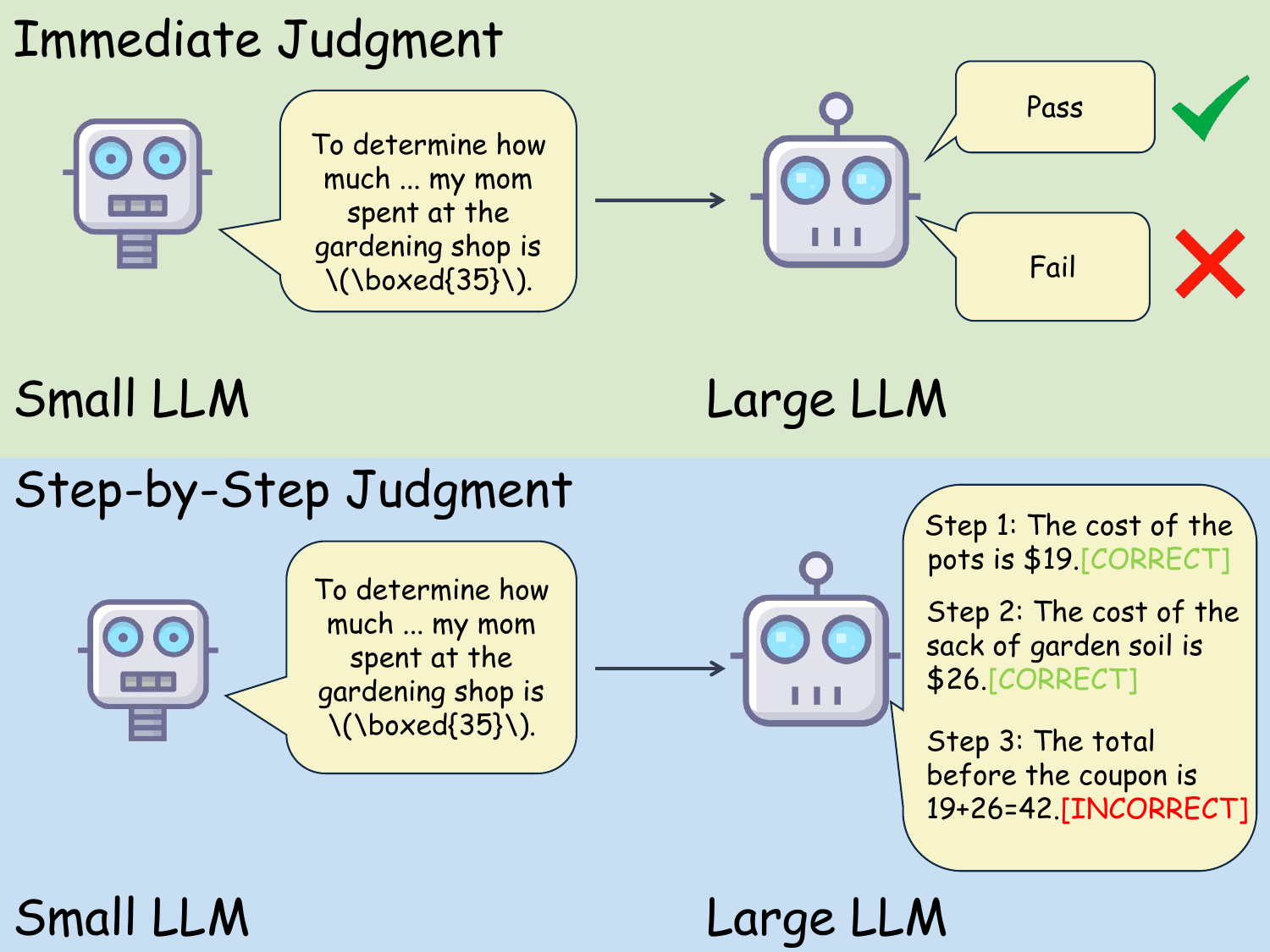}
    \end{subfigure}
    \caption{Two judgment strategies in our system. (1) Immediate judgment: the large LLM gives a direct judgment on whether the answer is correct. (2) Step-by-step judgment: Each step of the answer is labeled as [CORRECT] or [INCORRECT], and the process stops once an [INCORRECT] label appears.}
    \label{fig:2}
\end{figure}

\subsection{Pipeline}
Motivated by the challenges discussed above, we begin by presenting the agent-based pipeline of our approach, which is composed of one small LLM and one large LLM. Our pipeline is shown in Figure \ref{fig:1}. Upon receiving a question, the question is initially processed by the small LLM to generate a preliminary response. For clarity, we refer to the problem as \textit{$Q_t$} and its corresponding answer as \textit{$R_t$} throughout this work. Once the small LLM has generated a response, we concatenate $Q_t$ and $R_t$ to form a prompt denoted as $P_t$, which is then evaluated by the large LLM.
Following the evaluation, the model yields one of two possible outcomes for $R_t$: correct or incorrect. If the result is correct, the answer $R_t$ is directly returned. Otherwise, the question $Q_t$ is passed to the large LLM, which engages in deeper reasoning before generating the final answer.

\subsection{Judgment Mechanism}
This judgment stage represents the most critical component of the entire pipeline. We propose two evaluation strategies to verify the validity of the small LLM’s preliminary response, as shown in Figure \ref{fig:2}.

\textbf{Immediate Judgment.} The first strategy directly instructs the large LLM to assess the response without any intermediate reasoning. Specifically, the large LLM evaluates the combination of the given $P_t$ and produces a direct judgment of ``Pass'' or ``Fail''. If the evaluation result is ``Pass'', the small LLM's answer is directly adopted as the final output. If it is ``Fail'', the question is submitted to the large LLM once again for deeper reasoning before producing the final answer.

\begin{table*}[htbp]
\caption{Main experimental results of our proposed pipeline. (Immediate) denotes the immediate judgment strategy. (Step) denotes the step-by-step judgment strategy. Tokens denotes the number of model output tokens. Cost denotes the monetary expense for LLM API call (if applicable) of LLM per 100 question (USD).}
\label{1}
\centering
\begin{tabular}{l|ccc|ccc|ccc}
\hline \hline
\multirow{2}{*}{Model}& \multicolumn{3}{c|}{GSM8K} & \multicolumn{3}{c|}{MMLU} & \multicolumn{3}{c}{Avg} \\ 
\cline{2-10}& Accuracy & Tokens & Cost & Accuracy & Tokens & Cost & Accuracy& Tokens& Cost  \\
\hline
\multicolumn{10}{c}{\textit{Small LLM}}  \\ \hline
Qwen2.5-3B-Instruct &85.3& 330&0.004  &67.7 & 311&  0.004&76.5 &321 &0.004     \\
Llama3.2-3B-Instruct &76.9 &268 & - &54.8 &169 & - &65.9 &219 & -   \\
\hline
\multicolumn{10}{c}{\textit{Large LLM}}\\ \hline
Gemini-2.5-Pro &96.4 &1804 &1.804 &94.6 &1592 &1.592 &95.5 &1698 &1.698   \\
\hline
\multicolumn{10}{c}{\textit{Our Pipeline}}\\ \hline
Qwen+Gemini(Immediate) &93.5 &560 &0.297 &92.8 &893 &0.693 &93.2 &727 &0.495   \\
Qwen+Gemini(Step) &93.7 &795 &0.511 &93.0 &1257 &0.894 &93.4 &1026 &0.703   \\ 
Llama+Gemini(Immediate) &92.1 &597 &0.376 & 90.0&829 &0.671 & 91.1&713 & 0.524  \\
Llama+Gemini(Step) &92.5 &836 &0.521 &90.1 &1081 & 1.011&91.3 & 959&0.766   \\
\hline \hline
\end{tabular}
\label{tab:model_compare}
\end{table*}

\textbf{Step-by-Step Judgment.} The second evaluation strategy performs a step-by-step assessment of the response. For each input $(Q_t, R_t)$, $Q_t$ denotes the problem description and $R_t$ represents the small LLM’s preliminary response. We employ the large LLM to decompose $R_t$ into a sequence of explicit reasoning steps denoted as
\begin{equation}
r = \{s_1, s_2, \ldots, s_k\}.
\end{equation}
where $s_i$ refers to the $i_{th}$ individual reasoning step, and $k$ represents the total number of steps in the decomposed reasoning process.
After decomposition, the large LLM further evaluates the correctness of each step $s_i$ and appends a label as follows:

\begin{equation}
l_i \in \{\texttt{[CORRECT]}, \texttt{[INCORRECT]}\}
\end{equation}
If the current step is judged to be correct, the label \texttt{[CORRECT]} is attached to its end and the evaluation proceeds to the next step. Conversely, if the step is judged to be incorrect, the label \texttt{[INCORRECT]} is appended, and the evaluation process is immediately terminated. In this case, the result is returned to the large LLM for re-assessment through deeper reasoning. If all steps are deemed correct, the small LLM’s original answer is directly adopted as the final output.
As a result, we obtain the following labeled sequence:
\begin{equation}
\{(s_1, l_1), (s_2, l_2), \ldots, (s_t, l_t)\}
\end{equation}



\section{Experiments and analysis}
\subsection{Experimental Setup}

We conduct extensive experiments to validate the performance of our proposed system. The detailed experimental configuration is organized as follows.

\textbf{Models.} We adopt two open-source model families, Qwen and Llama, as our small LLMs. Specifically, we employ the Qwen2.5-3B-Instruct \cite{yang2025qwen3} model from the Qwen series, and the Llama-3.2-3B-Instruct \cite{dubey2024llama} model from the Llama series. For the large LLM with deep reasoning, we adopt Gemini-2.5-Pro \cite{comanici2025gemini} and enable its reasoning mode for all time.

\textbf{Datasets.} To evaluate the effectiveness of our approach, we select datasets with objective answers and relatively simple tasks. Specifically, we use the GSM8K \cite{cobbe2021training} dataset as one of the test sets, which contains 1,319 elementary-level mathematical word problems. In addition, we employ the MMLU \cite{hendrycks2020measuring} dataset (high\_school category), which consists of multiple-choice questions from three domains including physics, chemistry, and biology.

\textbf{Metrics.} For the evaluation, we use the commonly adopted \textit{accuracy} to measure the overall correctness of the pipeline, and the \textit{token count} to quantify the time efficiency of the entire reasoning chain. Since the token consumption differs between small and large LLMs (for example, API calls for large LLMs are more expensive), we also introduce a \textit{Cost} metric to compute the monetary expense incurred. \textit{Cost} represents the USD expense of LLM API calls (if applicable) per 100 questions.

\textbf{Baselines.} We directly evaluate the small and large LLMs on the test sets as a reference for our proposed method.

\subsection{Experimental Results}
As shown in Table~\ref{1}, when using small LLMs alone, the accuracy on both the GSM8K and MMLU datasets is already considerable. 
Specifically, Qwen2.5-3B-Instruct achieves an average accuracy of 76.5\% across the two datasets with only 321 tokens consumed. 
In contrast, although Gemini-2.5-Pro reaches an accuracy of 95.5\%, its token consumption increases to 1698, and the per-token cost of the large LLM is also more expensive. 
Our proposed method, however, reduces the overall consumption by half while sacrificing only about 2\% in average accuracy.
\begin{table}[htbp]
\caption{Extensive evaluation on more complex datasets. }
\label{2}
\centering
\begin{tabular}{l|cc}
\hline \hline
\multirow{2}{*}{Model}& \multicolumn{2}{c}{AIME2024}\\
 \cline{2-3}& Accuracy &  Cost \\
\hline
\multicolumn{3}{c}{\textit{Small LLM}}  \\
\hline
Qwen2.5-3B-Instruct &6.7&  0.01\\
Llama3.2-3B-Instruct &6.7  & -\\
\hline
\multicolumn{3}{c}{\textit{Large LLM}}\\
\hline
Gemini-2.5-Pro &93.3 &\textbf{15.32}  \\
\hline
\multicolumn{3}{c}{\textit{Our Pipeline}} \\
\hline
Qwen+Gemini(Immediate) &93.3  &15.33 \\
Qwen+Gemini(Step) &93.3  &15.63\\
Llama+Gemini(Immediate) &93.3 &\textbf{15.32}\\
Llama+Gemini(Step) &93.3 &15.54 \\
\hline \hline
\end{tabular}
\label{tab:model_compare}
\end{table}

\begin{table}[h]
\caption{Experimental results of incorporating verified correct steps into the prompt using Qwen with step-by-step judgment.}
\label{5}
\centering
\setlength\tabcolsep{5pt}
\begin{tabular}{l|cc|cc} 
\hline \hline
\multirow{2}{*}{Model}& 
\multicolumn{2}{c|}{AIME2024}&\multicolumn{2}{c}{GSM8K}\\

\cline{2-5} & Accuracy &  Cost  &Accuracy &  Cost \\
\hline
\multicolumn{5}{c}{\textit{Large LLM}}    \\
\hline
Gemini-2.5-Pro &93.3 &15.32 & 96.4&1.80  \\
\hline
\multicolumn{5}{c}{\textit{Our Pipeline}}   \\
\hline
w/o Correct &93.3  &15.63 &93.7 &\textbf{0.51}\\
w/ Correct &93.3  &\textbf{12.34} &93.2 &0.62 \\
\hline \hline
\end{tabular}
\label{tab:model_compare}
\end{table}

\begin{figure}[t]
\centering
    \begin{subfigure}[b]{1\columnwidth}
        \includegraphics[width=\linewidth]{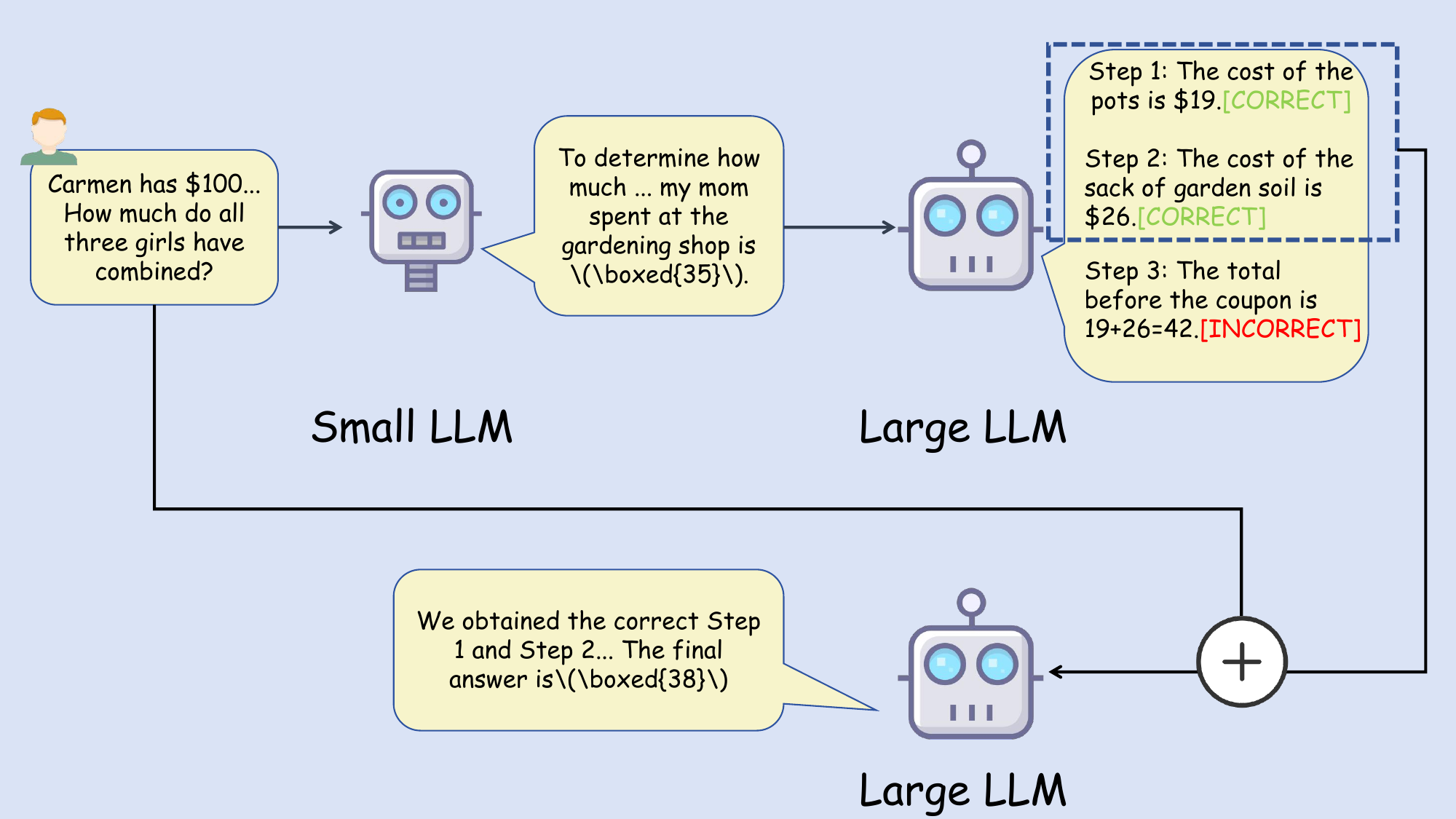}
    \end{subfigure}
    \caption{Incorporating verified correct steps into the prompt involves selecting only sentences labeled with [CORRECT], appending them after the question, and providing them to the large model for in-depth reasoning.}
    \label{fig:3}
\end{figure}


\textbf{Immediate Judgment vs Step-by-Step Judgment.} From the Table \ref{1}, it can be observed that the direct judgment strategy achieves slightly lower average accuracy compared to the step-by-step judgment strategy, while the latter incurs higher token consumption. 
This is because the step-by-step judgment evaluates each step of the small LLM’s response, making the evaluation more stringent. 
As a result, the error detection rate is higher, and more cases are handed over to the large LLM. 
Moreover, during the large LLM’s evaluation, additional tokens are consumed since it needs to output the judgment for each step. 
These two factors together explain why the step-by-step judgment strategy requires more tokens than the immediate judgment strategy. 
In terms of accuracy, however, the step-by-step approach is more rigorous, resulting in a lower false positive rate compared to direct judgment, and thus yielding higher overall accuracy.

\textbf{Evaluation on More Challenging Datasets.} We also conduct experiments on more challenging datasets. Specifically, we use AIME2024, on which the LLM judges the small model’s responses to be entirely incorrect.
As shown in Table~\ref{2}, the accuracy does not decrease, but the additional overhead introduced by the step-by-step judgment leads to higher overall consumption.
\\

\textbf{Incorporating Verified Reasoning into Prompts.}
In Table~\ref{2}, when facing complex problems, the step-by-step judgment incurs even higher overhead compared to the baseline. 
To alleviate this, we consider incorporating the correctly verified steps into the prompt to assist the large LLM in its deep reasoning, as illustrated in Table~\ref{5}. The pipeline is shown in Figure \ref{fig:3}.
After incorporating the verified steps, we conducted evaluations on both simple and complex datasets. 
We found that for simple datasets, adding the steps does not necessarily lead to more concise answers and may even increase the overall consumption. 
In addition, we observed a slight drop in accuracy when correct steps were provided for simple problems. 
In contrast, for complex problems, providing the correct steps reduces the consumption by nearly 20\% without harming accuracy. 

The underlying cause may be attributed to the behavior of deep reasoning models. Even when presented with correct intermediate steps, these models tend to re-evaluate them. For simple problems, the additional verification of already correct steps introduces computational overhead that exceeds the cost of the model’s own reasoning process, and this redundant reasoning may also introduce unnecessary errors, explaining the observed decline in accuracy. This observation aligns with recent studies, such as \cite{chiang2024over}, which demonstrate that excessive reasoning or redundant computation can impair performance on relatively simple tasks. Conversely, for complex problems, large LLMs typically engage in extensive self-reflection over their reasoning steps, so the cost of re-verifying externally provided correct steps is lower than that of generating and reflecting on their own reasoning. Consequently, this approach reduces the overall computational burden while maintaining accuracy.


\section{conclusion}
In this work, we proposed an efficient agent system that leverages the collaboration between small and large LLMs. By dynamically allocating tasks, our approach effectively reduces the unnecessary overhead caused by employing deep reasoning models on simple problems, while still maintaining strong performance on more complex tasks. This demonstrates a practical pathway toward building cost-efficient and accurate multi-model reasoning systems. In addition, we introduced two alternative judgment strategies and analyzed their respective strengths and limitations. This comparative analysis provides deeper insights into when each strategy is most effective, further guiding the design of efficient reasoning pipelines. We plan to explore dynamic methods for determining when to provide verified steps to the model in the future, so as to further optimize efficiency on complex tasks while minimizing unnecessary overhead on simpler ones.

\bibliographystyle{IEEEbib}
\bibliography{strings,refs}

\end{document}